\documentclass[conference]{IEEEtran}
\usepackage{amsmath}
\usepackage{amssymb}
\usepackage{graphicx}
\usepackage{url}

\DeclareMathOperator*{\argmin}{argmin}

\begin{document}

\title{Referential Uncertainty and Word Learning in High-dimensional, Continuous Meaning Spaces}

\author{\IEEEauthorblockN{Michael Spranger}
\IEEEauthorblockA{Sony Computer Science Laboratories Inc., Tokyo, Japan\\
Email: michael.spranger@gmail.com}
\and
\IEEEauthorblockN{Katrien Beuls}
\IEEEauthorblockA{Vrije Universiteit Brussel, Brussels, Belgium\\
Email: katrien@ai.vub.ac.be}
}

\maketitle

\begin{abstract}
This paper discusses lexicon word learning in high-dimensional
meaning spaces from the viewpoint of referential uncertainty. We investigate 
various state-of-the-art Machine Learning algorithms and discuss the impact 
of scaling, representation and meaning space structure. We demonstrate that
current Machine Learning techniques successfully deal
with high-dimensional meaning spaces. In particular, we show that exponentially 
increasing dimensions linearly impact learner performance and
that referential uncertainty from word sensitivity has no impact.
\end{abstract}

\IEEEpeerreviewmaketitle

\section{Introduction}
The most important aspect of word learning is often thought to be \emph{referential uncertainty}.
Quine \cite{quine2013word} famously framed referential uncertainty as a general problem
everybody faces when trying to learn an unknown language.
Suppose an anthropologist studies an isolated tribe. When members of the tribe see a rabbit,
they shout ``gavagai''. The anthropologist hears the word for the first time and has no idea what
the word means. In principle the meaning of the word could be anything 
from perceptual features of objects present (or not present!), features of the environment, social and
historical facts etc. The space of possible meanings is essentially infinite and
the question is how the anthropologist can solve this puzzle.

\begin{figure}
\begin{center}
\includegraphics[width=\columnwidth]{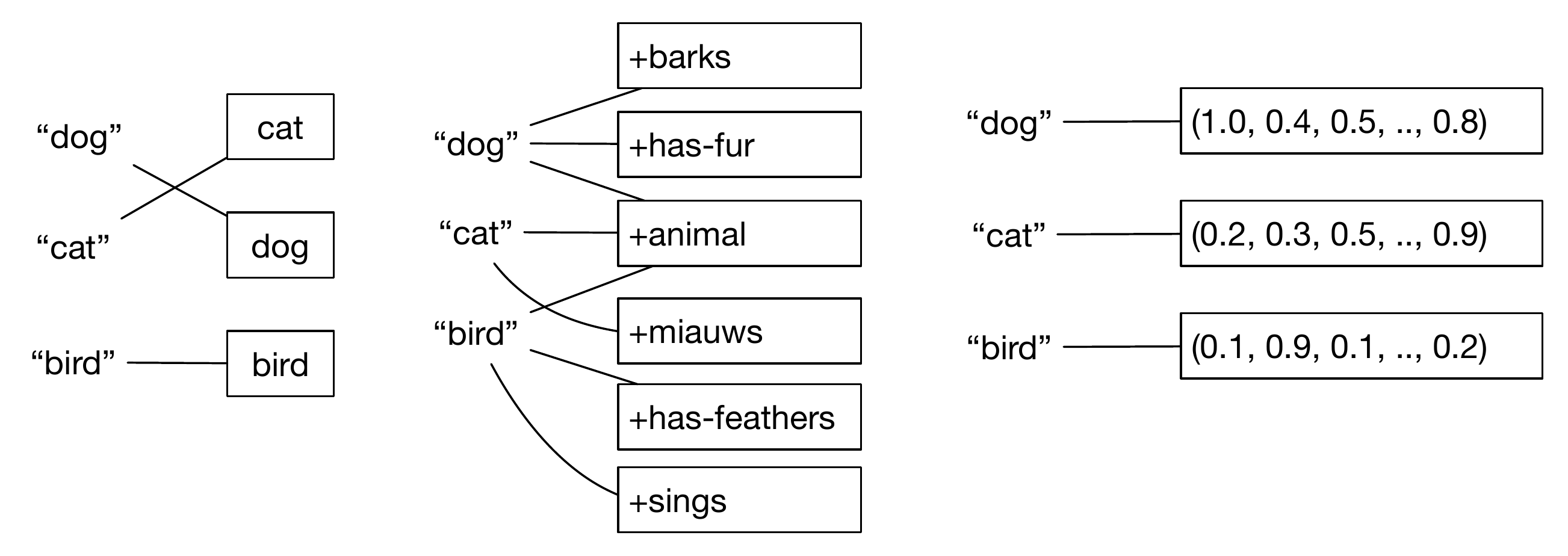}
\end{center}
\caption{Left: models of word-object mappings learn associations between words 
and a priori known objects.
Middle: symbolic feature based models learn
associations to symbolic feature representations. Right: word meanings 
are organizations of $n$ dimensional, continuous feature vectors.}
\label{f:models}
\end{figure}

Many researchers claim that children face the same problem
when trying to learn language. Models that deal with word learning try to capture 
referential uncertainty in various ways. Schematically, we can distinguish between 
the following approaches to meaning and corresponding models (see Fig. \ref{f:models}).

\paragraph{Word-Object Mapping Models (WOM)}
Some models approach word learning as a discrete mapping problem from 
words to object, e.g. \cite{fontanari2009cross,kachergis2012associative}. How difficult the mapping problem is depends 
on factors such as the number of objects in the context of an interaction or the feedback given by 
the caregiver. If there is only one object, then the mapping can be learned instantaneously and 
no problem exists. If there are multiple objects, then referential uncertainty does exist if 
there is ambiguous, unreliable or no feedback from the caregiver. 

\paragraph{Combinations of Feature Models (CFM)}
Some researchers \cite{siskind1996computational,deBeule2006cross-situational,wellens2008coping} 
have modeled word meaning as combinations of symbolic features. The features themselves 
are known prior to word learning.  In CFM the meaning space is as large as all possible 
combinations of features. The meaning space grows combinatorially with the number of features.
There are fundamental differences between WOM and CFM. In CFM, a word can 
be linked to multiple features. The learner upon hearing a word and seeing a single object cannot know
which features of the object the word refers to. Consequently, referential uncertainty 
can occur in single object contexts  This is different from WOM, where referential uncertainty 
is exclusively related to the number of objects in the context.

\paragraph{Continuous Meaning Space Models (CMS)}
Few models address the learning of words related to representations in 
continuous vector spaces \cite{belpaeme2012acs,spranger2013acquisition}. 
The problem of referential uncertainty in continuous meaning spaces is large and 
depends chiefly on the number of objects in a learning
context, the number of dimensions of meaning vectors and whether
or not objects can refer to subspaces of the meaning space (color space etc).
In that sense CMS behave similar to CFM. The difference to CFM is that
the meaning space is an infinite continuous vector space by default.

The few available CMS models are often shown to work in low-dimensions 
and with few words. In this paper 
we survey the state of the art in word learning algorithms 
for high-dimensional continuous meaning spaces with referential uncertainty 
caused by words referring to features of objects (rather than caused by multiple objects in the context).
As it turns out, this problem has been dealt with in the Machine Learning (ML)
community. We first analyze the problem of word learning
from the viewpoint of machine learning in $n$-dimensional spaces. 
We then test various state-of-the-art ML methods on simulated and grounded 
data sets, analyze dynamics of online learning and the impact of various 
referential uncertainty. Lastly, we discuss
how all this fits in the general landscape of language learning models.

\section{Description Games}
\label{s:tutor}
One interaction pattern (game) often used in models of word learning is the 
\emph{description game} (DG). The basic structure of DG is the following.
The learner observes a particular situation (context) of $l$ objects ($l=1$ for this paper) 
and also observes $k$ words uttered by the tutor ($k=5$ for this paper). So for instance, 
both learner and tutor observe one object. The tutor says ``block, bright, red,..,..'' (order 
does not matter). The learner then has to learn the meaning of 
these words by integrating information over various trials. 
The success of the learner is measured by testing which words the learner produces for 
objects and how this overlaps with the production of words of the tutor.
 
An important aspect of these games is the \emph{tutor strategy} - the 
representation and algorithm the tutor uses for producing $k$ words for an object. 
In this paper, objects are represented by $n$ dimensional feature vectors $o\in[0,1]^{n}$.
The tutor represents each word using a prototype $p\in[0,1]^{n}$ and a weight vector $w\in[0,1]^{n}$. Prototypes 
for tutors are randomly drawn from a uniform distribution $\mathrm{U}(0,1)^n$. 
Weights are drawn from a binomial distribution $\mathrm{B}(1,0.5)^n$. If a weight 
vector is all 0 we randomly set one of the weights to 1. 

For an object $o\in[0,1]^{n}$ the tutor first computes a weighted Euclidean distance $\operatorname{wd}$ 

$$\operatorname{wd}_{w,p}(o) = \sqrt{ \sum_{i = 1}^{n} w_i (o_i - p_i)^2}$$

The tutor then chooses the $k$ closest words for the description of an
object, i.e.
$$\argmin_{w,p \in P}(\operatorname{wd}_{w,p}(o))$$

This representation and word production strategy has the following desirable properties. 
The strategy models feature dimension sensitivity of words/prototypes. 
For instance, there could be a word that is sensitive only to the brightness
dimension. Other words could be sensitive 
to blue and red channels. Secondly, words are not produced uniformly - similar to 
human language.  Some words are used often, others only occur 
few times in the training set. Lastly, the tutor produces $k$ words for any object. Together 
with the object distribution in $[0,1]^n$ this leads to interesting, non-linear interactions 
between words and objects.

Let us briefly examine the difficulty of the learning task. 
The learner has to learn to produce (predict) the same words 
as the tutor based on examples provided by the tutor.
Suppose that $m=|W|$ denotes the number of words the tutor knows. 
In principle, chance performance is equal to choosing $k$ out of $m$ 
words without repetition and order does not matter. For experiments 
with $k=5$ and $m=100$ this amounts to ${100 \choose 5}  = 75,287,520$
assuming a uniform distribution of words. 

\section{Description Games and Machine Learning}
\label{s:learner}
In machine learning terms, DG is a \emph{supervised}, 
\emph{multi-class} (multiple words), \emph{multi-label} (multiple words per object), 
\emph{online} classification problem. 
There is an immense literature and an abundance of algorithms that can be tested on 
this problem \cite{zhang2014review,tsoumakas2009mining}. 

For this paper, we test roughly a dozen different learning methods from linear models,
to ensemble classifiers, Bayesian learning and neural networks that solve the description 
game learning problem. The following paragraphs give a (brief) overview of the various 
methods we tested. 

\paragraph{Nearest neighbor (NN)}
One of the simplest and often best performing methods is nearest neighbor -
also called KNN or in this paper \emph{KNeighbors} \cite{cover1967nearest} .
KNeighbors is a \emph{non-parametric} method that stores all samples
ever encountered. New samples are classified based on the class of its $k$ 
nearest neighbor (stored examples). 
The algorithms simplicity and the (often out of the box) success of this method have led 
to its widespread adoption. 

We also use a related algorithm from the 
same family: \emph{Nearest Centroid (NC)}. NC represents classes 
using centroids of corresponding samples. New samples are classified
based on the shortest distance to centroids. This method is among 
the most widely used in word learning because it corresponds nicely
with ideas in psychology \cite{rosch1975cognitive}. 

\paragraph{Generalized Linear Models (GLM)}
GLM describe a family of algorithms that all model predictions
as linear combinations of input variables. All models furthermore assume	
that predictions (dependent variables) are generated from exponential 
probability distributions (Gaussian, binomial, gamma etc). Learners also differ in 
terms of regularization and learning regime (closed-form, stochastic etc). We are
using various classifiers: \emph{Logistic Regression}, 
\emph{Online Passive Aggressive} (PA) \cite{crammer2006online} and 
\emph{Stochastic Gradient Descent} (SGD) \cite{bottou2010large}.

\paragraph{Ensemble methods (ENM)}
ENM are meta algorithms that try to improve classification results by combining results
of sets (ensembles) of classifiers (NN, linear or others). There are basically
two types of ensemble methods. The first relies on weak, underfitting classifiers each 
often only slightly better than random choice. There are two main methods in this group: \emph{AdaBoost} \cite{freund1995desicion} 
and \emph{Gradient Boosting} \cite{friedman2001greedy}. AdaBoost works by 
fitting a series of weak learners.
Each step (boosting iteration), training data is weighted to focus on samples that
are not correctly classified in the previous step. Successive classifiers therefore
essentially encode classification for various aspects of the data. New samples are 
classified by computing the majority estimates of classifiers.
Gradient Boosting is an extension of boosting for optimizing any differentiable loss function.

Another class of ensemble methods takes the opposite
approach and relies on sets of classifiers
that are complex, over-fitting classifiers. Here, we use \emph{Random Forest} \cite{breiman2001random} 
and \emph{Extra Trees} \cite{geurts2006extremely}, which use
ensembles of decision trees. Decision trees are a non-parametric method that 
learns binary decisions (nodes) and arranges them in a binary tree 
\cite{rokach2014data}. Both RandomForest and ExtraTrees 
build multiple overfitting classifiers on random subsets of samples and features.

\paragraph{Bayesian Methods (BM)}
BM rely on Bayes theorem to transform
the classification problem into one of estimating 
probability distributions. New samples are classified
based on prior probabilities of classes, as well as 
posterior estimates of sample probabilities and 
the probability of observing samples given classes.
Bayesian classifiers primarily differ in the assumptions they make 
for the probability distributions that need to be estimated. 
We use \emph{Gaussian Naive Bayes} 
classifiers \cite{jordan2002discriminative}
(normal distributions, independent features)
and \emph{Multinomial Naive Bayes} (multinomial distributions, independent features).
Parameters for probability distributions are 
estimated using expectation-maximization.

\paragraph{Neural networks (NN)}
Recently neural networks have pushed the state-of-the-art in many
classification problems such as face recognition, image labeling
etc. The current trend is to stack multiple layers of neurons (mostly non-linear
functions) and train them (often one layer at a time) using variants of 
the backpropagation algorithm. NN can take various forms
in terms of network topologies, transfer functions, training regimes and 
learning rules. For the purpose of this paper, we used
a multi-layer perceptron (\emph{MLP})  - (2 layers, rectified linear units, with a final 
sigmoid layer for classification).

\begin{figure}
\begin{center}
\includegraphics[width=.9\columnwidth]{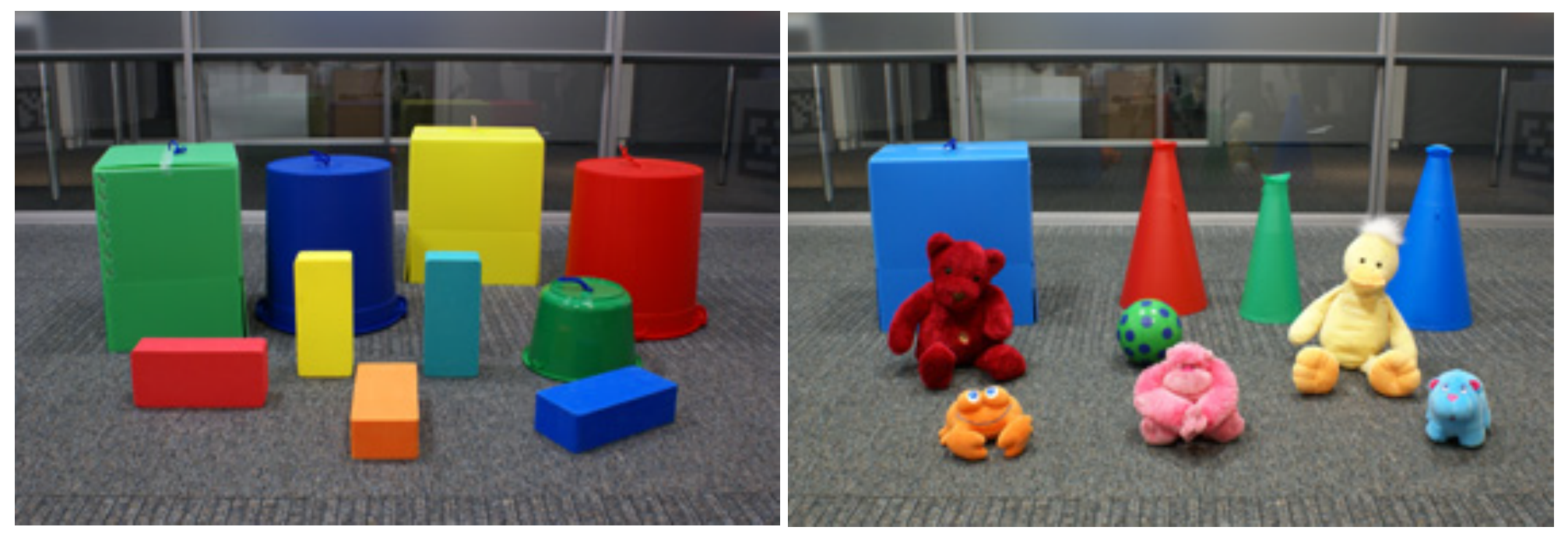}
\end{center}
\caption{Objects used in evaluating different algorithms.}
\label{f:objects}
\end{figure}

\section{Experimental Setup}
\subsection{Datasets}
We compare different learners on 
simulated and robot data sets. The robot data sets consists of 20 
different objects (see Figure \ref{f:objects}) of various color, shape, size.
We recorded approximately 1000 scenes in which two robots observe objects
from different perspectives. Scenes differ in the positions of objects, which
also alters objects' perceived shape and color features.
For each object $n=17$ feature dimensions are extracted: x, y, z position, YUV color values 
(mean min, max), width, height, length. Data is scaled between 0 and 1 using a linear scaler
and for some classifiers to zero mean, unit variance.

\paragraph{GRO1} This dataset consists of all object observations ever made by the 
two robots in one matrix ($4532 \times 17$). GRO1 is used to experiment, with tutor 
and learner having the same object perception, in particular,
the same feature estimation for objects. 
\paragraph{GRO2} This dataset is a grounded robot data set. It consists of two matrices. One
for each robot (two matrices of size $4532 \times 17$). Each row corresponds
to the perception of the same physical object from the viewpoint of the two different robots.
This data is used in the following way. The tutor produces words from the perspective
of robot 1 (matrix 1). The learner learns from the observation of the same object but from
perspective of the other robot (matrix 2). \emph{GRO2} is used to evaluate what happens if there 
is \emph{perceptual deviation} \cite{spranger2012deviation}.
That is tutor and learner see the scene from different viewpoints and 
therefore have different feature estimations for objects. For instance, the tutor robot
might observe different x, y positions for an object, since he sees the
object from a different perspective. But also color and shape features
will be slightly different.
\paragraph{SIM} This dataset is simulated and consists of 4532 object
observations of $n=17$ feature dimensions drawn from a uniform 
distribution $\mathrm{U}(0,1)$.

\subsection{Methods}
Learners are trained on samples of objects and one-hot vector
encoded words produced by the tutor. Each classifier has to predict 
(produce) the correct set of $k$ words given (new)
vectors of objects by predicting a one-hot vector encoding of word
activations. We then measure the difference between the production
of the tutor and the prediction of the learner. For learning algorithms
that predict probabilities $p$  for words (e.g. MLP), if $p>0.5$
then the word is counted as a prediction.

For all experiments here, we draw $|W|=100$ prototypes and 
weights for the tutor (according to the description in Section \ref{s:tutor})
and perform 4-fold validation on data sets each consisting of 4532 samples. 
This means that training happens on roughly 3400 samples and testing on 
1100 new samples. In summary, the standard parameters for our evaluations 
are $n=17$ (number of features), $|W|=100$ (number of words 
and prototypes tutor), $k=5$ (number of words uttered by the tutor)
and $p=0.5$ (for the binomial distribution of tutor weights).

Classifiers not supporting multi-class, multi-label by default were trained using one-vs-rest \cite{spolaor2015systematic}.
The exception are RandomForest, 
KNeighbors, NearestCentroid and MLP. 
Most classifiers rely on various hyper-parameters. We optimized 
hyper-parameters using parameter grid searches on a separate simulated dataset 
\emph{SIM-DEVELOP} (same characteristics as \emph{SIM}).
Hyper-parameters were optimized once on \emph{SIM-DEVELOP} and 
then fixed for all results reported here. 

\subsection{Measures}
In this paper we use a single performance measure: \emph{f-score}
\cite{sokolova2009systematic}. F-score is defined as the harmonic mean of \emph{precision} 
and \emph{recall}. There are various definitions of precision and recall depending
on the classification problem (binary, multi-class, multi-label). Generally 
speaking, precision measures the amount of wrong words per sample
found by the learner. Recall measures how many of the correct words
were predicted by the learner. We use a particular f-score 
measure which is called \emph{example-based} (or \emph{sample})
that does not take into account unbalanced word distributions.
An f-score of 100 means that all words and only those words uttered by 
the tutor are uttered by the learner. 

\section{Results}
\begin{table}[t]
\begin{center}
\begin{tabular}{| l | c | c | c |}
\hline
 & SIM & GRO1 & GRO2 \\\hline
 \emph{Nearest Neighbor} & & & \\\hline
KNeighbors & 67.89 & 74.06 & 61.25\\\hline

NearestCentroidOvR & 65.81 & 83.44 & 63.87\\\hline

\emph{Linear Models} & & & \\\hline
SGD & 79.49 & 87.93 & 65.27\\\hline

PassiveAggressive & 63.32 & 84.70 & 65.04\\\hline

LogisticRegression & 76.61 & 86.17 & 61.95\\\hline

\emph{Ensemble Methods} & & & \\\hline
RandomForest & 80.84 & 87.91 & 65.79\\\hline

ExtraTrees & 67.89 & 74.06 & 61.25\\\hline

AdaBoost & 79.49 & 87.93 & 65.27\\\hline

GradientBoosting & 80.83 & {\bf 92.58} & 65.68\\\hline

\emph{Bayesian} & & & \\\hline
GaussianNB & 63.85 & 84.44 & 65.25\\\hline
MultinomialNB & 75.35 & 87.02 & 67.64\\\hline
\emph{Neural} & & & \\\hline
MLP & {\bf 82.74} & 91.58 & {\bf 70.87} \\\hline
\end{tabular}
\end{center}
\caption{Results comparison grounded and simulated data (sample-based f-score, $n=17$, $k=5$, $p=0.5$, $|W|=100$)}
\label{t:experiment-1}
\end{table}

Table \ref{t:experiment-1} shows the performance of various classifiers on
grounded and simulated data.
Many learners perform well on the task with respect to task complexity. 
In particular, simple algorithms such as GaussianNB or linear models perform 
well. More complex methods such as ensemble
methods generally are top performers. The best performing method 
on \emph{SIM} and \emph{GRO2} is the multi-layer perceptron MLP. 
On \emph{GRO1} GradientBoosting is the front runner. Although
ensemble methods generally perform quite similar. None of the methods
fail catastrophically, which is mostly due to hyper-parameter optimization. 

Interestingly all methods improve on grounded data 
(\emph{GRO1}) some by as much as 20 points (e.g. PassiveAgressive). This
suggests that methods are able to take advantage of structure available in grounded data.
However, all methods perform worse on \emph{GRO2} than on \emph{GRO1}.
In some cases performance differs by almost 30 points between
\emph{GRO2} and \emph{GRO1}.

A word on how to understand these results. This study focusses on understanding
the baseline for word learning in high-dimensional meaning spaces. These
results are existence proofs (lower bounds) showing that learners can solve this problem with relatively 
high f-score. These results do not show limits of individual learning algorithms, rather
results show general trends that hint at the scope of the learning problem.

\subsection{Scaling Object Feature Dimensions}
\begin{figure}
\begin{center}
\includegraphics[width=\columnwidth]{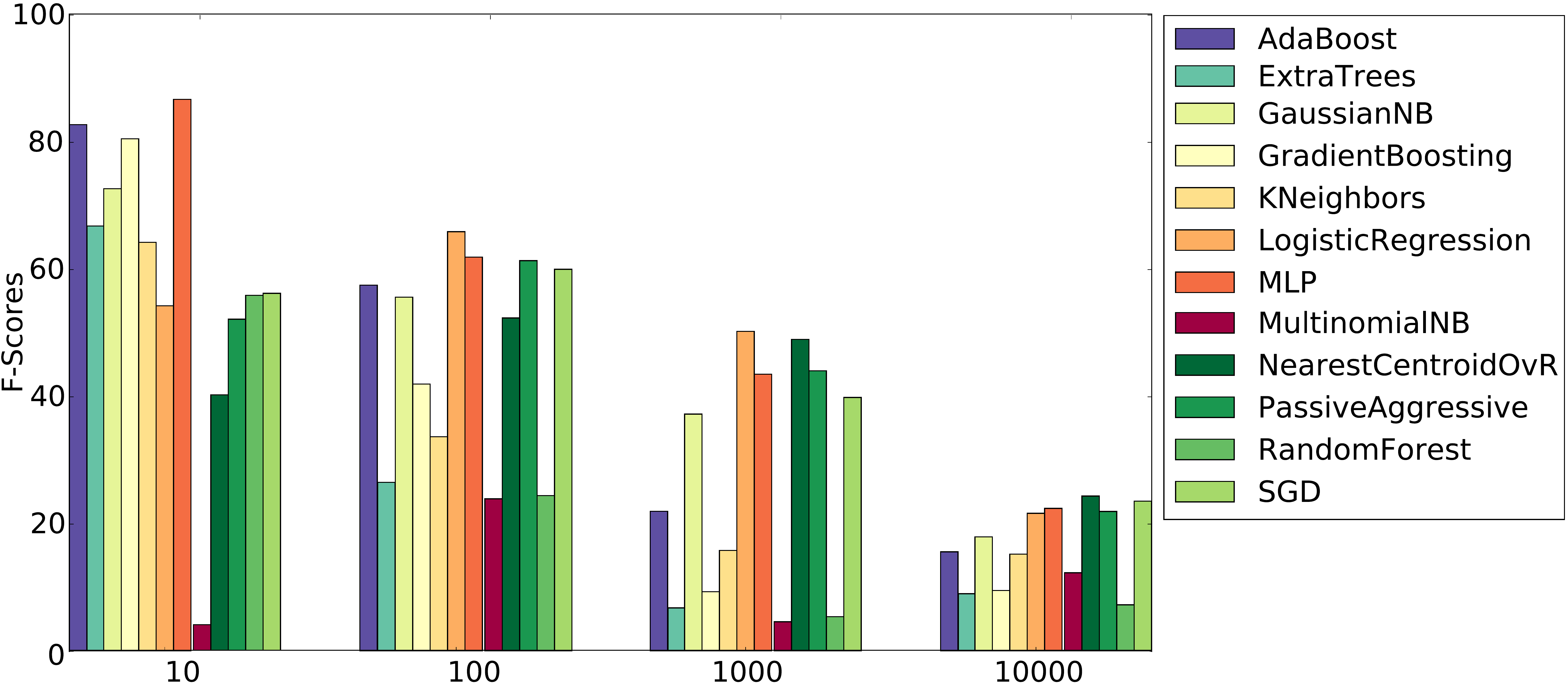}
\end{center}
\caption{Performance of classifiers for increasing $n$ (number of object feature dimensions) 
with $p=0.5$, $|W|=100$, $k=5$, data set size 4532, 4 fold.}
\label{f:scaling}
\end{figure}

We are interested in understanding referential uncertainty in 
high-dimensional meaning spaces. Consequently, we manipulated 
the number of dimensions of object features $n$. The number 
of dimensions in the grounded data sets is fixed by the vision system, so we 
increased the number of dimensions for simulated data.
All other parameters are kept the same. 
Figure \ref{f:scaling} shows how classifiers do for $n \in \{10, 100, 1000, 10000\}$. 

Average performance across all classifiers degrades linearly with orders of magnitude of 
difference in $n$ for the best performing classifiers (MLP, AdaBoost).
These results suggest that classifiers optimized for various $n$ and/or increasing 
the number of training samples, could actually 
deal with even higher $n$-dimensional data (remember that 
all classifiers were optimized for $n=17$). 

There is one classifier that performs poorly (MultinomialNB) all along, while
others (e.g. ExtraTrees, RandomForest) degrade much more rapidly with number 
of dimensions than the best performing ones. Others perform
best for certain $n$ (e.g. PassiveAggressive). All of these classifiers perform 
reasonably well or very well on the hyper-parameter optimized $n=17$. 
Consequently, these classifiers are sensitive to hyper-parameters with respect to 
number of dimensions. 

\subsection{Scaling Word Sensitivity}
Another dimension of scaling is the sensitivity of words. Prototype weights
for all experiments reported so far were drawn from a binomial distribution $\mathrm{B}(1,p=0.5)$. 
This means that words are on average sensitive to half of the dimensions. 
In our experiments, referential uncertainty is tied to the fact that words can refer to 
aspects of objects. To test learners, we ran experiments for 
various $p\in \{0.1, 0.25, 0.5, 0.75, 1.0\}$ and $n=100$. All other parameters stay the same.

\begin{figure}
\begin{center}
\includegraphics[width=\columnwidth]{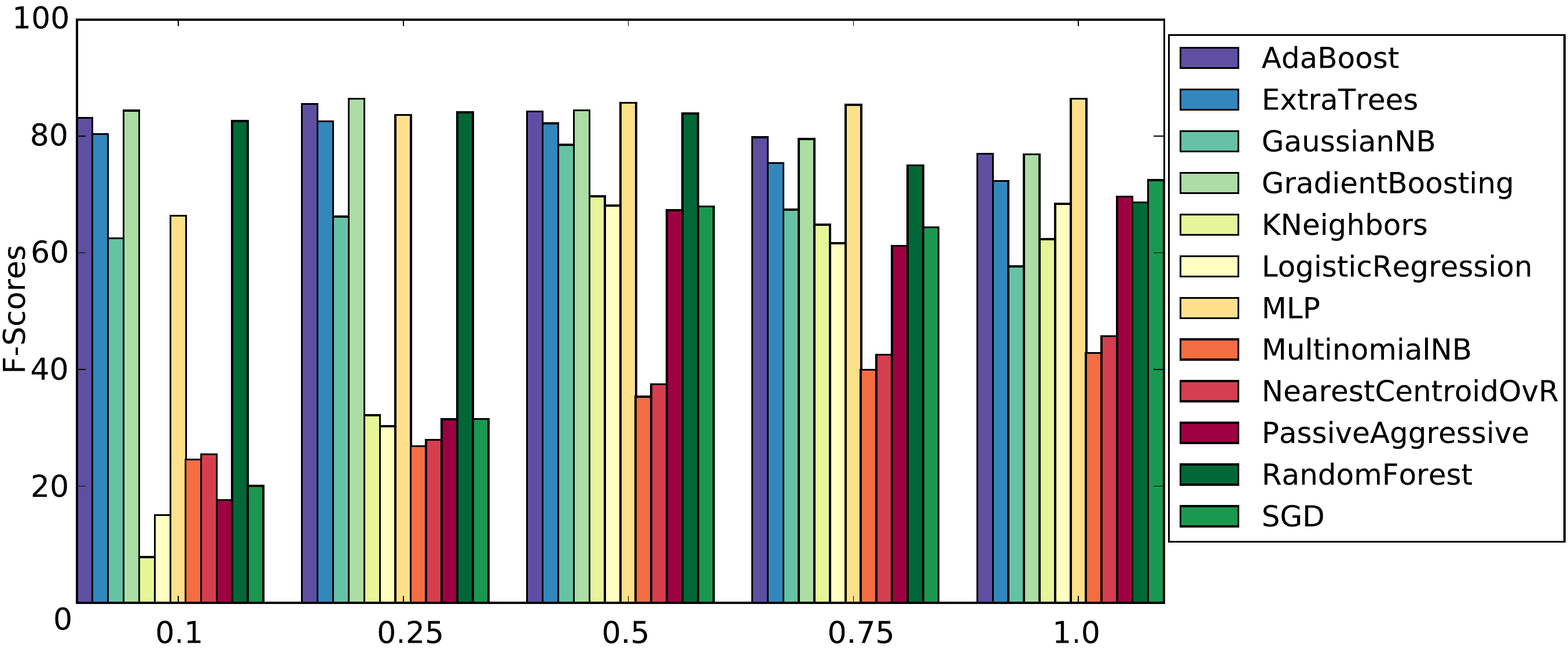}
\end{center}
\caption{Performance of learners for varying values of $p \in\{0.1,0.25,0.5,0.75,1.0\}$ with
$n=100$, $|W|=100$ and $k=5$.}
\label{f:scaling-p}
\end{figure}

Figure \ref{f:scaling-p} shows that quite a few learners (AdaBoost, ExtraTrees, MLP etc) have
absolutely no problem dealing with various $p$. We can conclude that these learners
will perform well on mixed languages where some words encompass all features of an object
and some are more specific and only refer to certain features. Other learners such as KNeighbors
become better with larger $p$. This is no surprise since KNeighbors stores full examples.

\subsection{Online Learning}
\begin{figure}
\begin{center}
\includegraphics[width=0.8\columnwidth]{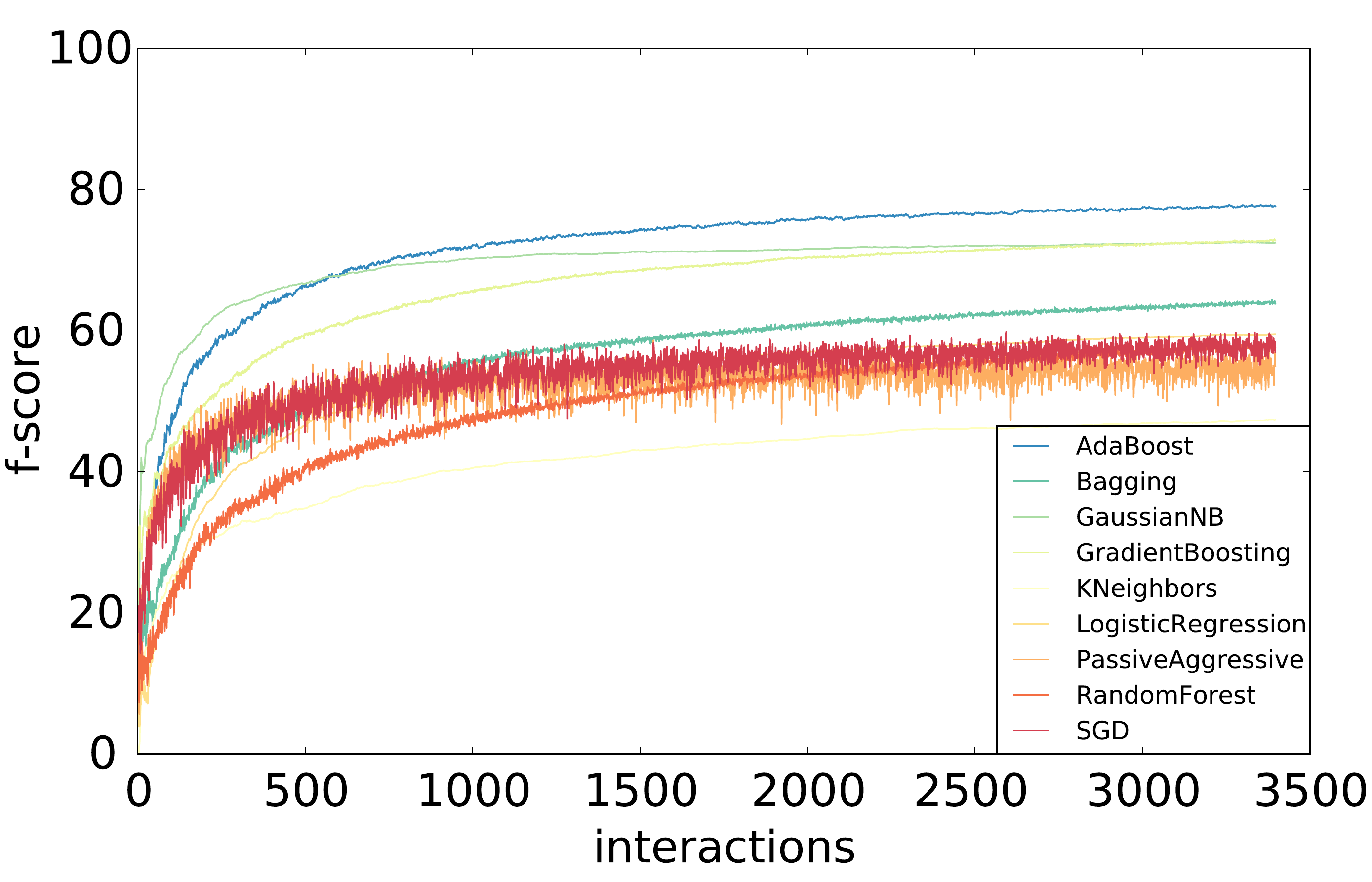}
\end{center}
\caption{Online learning performance of various classifiers on simulated data.}
\label{f:online}
\end{figure}

One important aspect in language learning from a developmental
point of view is online learning. We tested how well the algorithms 
perform over time. For this we incrementally train classifiers
on the training set. For instance, we train on the first $m$ object observations 
and evaluate the f-score on the test data set. Figure \ref{f:online}
shows the performance of classifiers over time. Pretty much
all classifiers learn very fast with most of the gains happening
on the first 500 training samples. In other words, the learner
becomes quite sufficient after 500 interactions with the tutor. 
After 1000 training samples all classifiers are within 5 points 
of their final f-score.

These results for online learning are quite remarkable given that 
there are 100 words that need to be learned. Certainly, what helps 
here is that certain words will be used frequently and others less frequently.
We analyzed results with respect to frequency of words
and it becomes evident that this is indeed one big driver
of the speed of learning.

\subsection{Discussion of Results}
Our results show that many off-the-shelf machine learning algorithms 
can deal with high-dimensional meaning spaces. 
We only optimized classifiers for $n=17$ and we get linear
drop in performance for exponential increase in number of dimensions. 
This suggest that we might be able to get on par performance
even for very high $n$ - if we optimize classifiers for each $n$
and if we logarithmically increase training set size. Our results also show that 
many classifiers have no problem dealing with feature sensitivity related to 
referential uncertainty.

Our experiments suggest that referential uncertainty in high-dimensional 
meaning spaces is NOT an exponentially growing problem. In other
words adding dimensions has at most a linear effect on learners in the paradigm
discussed here. This is an interesting result because it suggests
that referential uncertainty although often thought of as an exponentially
growing problem with the number of dimensions or the degree with which words 
can associate to aspects of objects is much less of an issue than one might expect.

\paragraph{Meaning Space Structure}
There is a difference in performance between \emph{SIM} 
and \emph{GRO1} data sets. The reason is that 
there is much more structure in the real world than
in the simulated world. The real world data sets consist
of limited sets of objects that make up very defined spaces in the
perceptual data space. For instance, there are certain clearly separable
color regions based on the objects in \emph{GRO1}.
This structure in the environment helps all learning algorithms 
in becoming more successful.

\paragraph{Tutoring}
Tutoring strategies are often thought to be about social feedback
(e.g. pointing or agreement). But tutoring can also mean that the tutor is structuring the 
environment (and possibly also the language) 
for the learner. This can include taking perspective or conceptualizing the world
from the viewpoint of the learner. Generally speaking it has been found that 
tutoring strategies help learners \cite{spranger2013acquisition,spranger2015incremental}.
The delta in performance between \emph{GRO1} and \emph{GRO2} confirms
these ideas. In \emph{GRO1} the tutor will utter words based on
what the learner sees. In \emph{GRO2} that is not the case. 
All classifiers perform less well on \emph{GRO2}.

\paragraph{Unbalanced data}
Another aspect that affects performance is the 
fact that training of words is unbalanced. There are 
some words that occur often and others that don't
(in fact some words do not even appear in the training set
and only in the test set). The classifiers have  
difficulties with sparsely used words. Something that 
becomes apparent when examining macro-averaged 
f-scores (not reported here). This is often much lower 
than micro-averaged f-scores. This split suggests that
(generally speaking) learners are good in learning 
frequent words but less good in learning less frequent words.

\paragraph{Representation}
It is interesting to analyze various learning algorithms with respect
to whether they actually build representations similar to that of the tutor.
We deliberately chose various algorithms none of which directly 
tried to replicate the tutor behavior in the learner by learning the same 
representation. The tutor operates using weighted feature distances
to prototypes. Words are only sensitive to a particular feature channel (e.g. the brightness).
Algorithms such as KNeighbors do not explicitly represent
information like that. They just collect samples. Others such as RandomForrest 
do actually learn how to distinguish different words based on explicitly
learning which features matter with respect to the word. An interesting 
result of this study is that both of these algorithms perform comparably 
well in terms of replicating the tutors behavior. But if we look at different
$p$ value experiments, we can see that discriminative feature learners
such as AdaBoost outperform KNeighbors.

\section{Related Work}
\paragraph{Referential uncertainty} 
There is an important difference between the setup here and other studies. Many studies are
concerned with enumerable objects in context and how this leads to referential uncertainty (see \cite{frank2008bayesian,belpaeme2012acs,deBeule2006cross-situational} among others). 
In this paper, we use referential uncertainty closer to Quine's formulation and early studies by Siskind \cite{siskind1996computational}. 
Quine focusses on aspects of a situation and not on enumerable objects as the source of referential uncertainty.
In that sense the problem in Quine is larger. Even if you know the referent, the learner
still knows nothing about the aspect of the referent the word refers to (color, shape etc).
The question remains which referential uncertainty problem is solved by children (possibly both).

\paragraph{Description vs Discrimination} An important distinction between
various models is the tutoring strategy - representation and algorithm for 
word production in the tutor. In description games, the speaker
minimize distances between the topic object and words (here weighted Euclidean distances).
In other types of interaction (called \emph{guessing games}),
the function being maximized for each word is the 
difference between the topic object $t$ and all the other objects (or features)
in the context.  An interesting question is whether different production strategies require
different learning algorithms or not. Often the learner is modeled
after the tutor and both use the same production and interpretation algorithms 
(see \cite{spranger2013acquisition,wellens2008coping} for some recent examples). 
This obviously biases the system and the question is whether this is necessary. 
What we can say for the models described in this paper is that nowhere did we bias 
the learners explicitly towards a particular production algorithm. Rather all that happens 
is that the learner is trying to replicate the tutor behavior.

\paragraph{Child Learning Strategies}
Researchers in child language acquisition have provided many ideas
about strategies that children use to learn the meaning of words. Some of them such
as \emph{perceptual biases} \cite{pruden2006birth} could potentially be exploited
by learning algorithms - if the language affords it. That is, 
if the language to be learned is based on salient perceptual distinctions then
algorithms that learn discriminative features (e.g. decision trees, ensemble 
methods) can take advantage of that. Other child language learning strategies 
based on \emph{linguistic constraints} \cite{gleitman1990structural} 
are not by definition part of the learning paradigm discussed in this paper. We
only focus on unordered sets of words uttered by the tutor.
The impact of strategies such as \emph{mutual exclusivity} \cite{markman1988children} 
and \emph{contrast} \cite{clark1987principle} on the learning problem
defined in this paper is subject to future work.

\section{Conclusion}
Abstract models can be used to answer questions about what algorithms 
children use to learn language. This is often done by trying to 
replicate empirical data.  Another goal of models is to grasp 
the essence of the learning problem, characterize how hard it is and
how the best known algorithms perform. This paper addresses
aspects of the second goal. We defined an abstract version of 
the word learning problem, translated it to Machine Learning and 
compared state of the art methods on the problem. This establishes
a baseline that can be used to understand word learning
from the viewpoint of complexity. 

Source code as well as data sets are published online at \url{https://github.com/mspranger/icdl2016language}.

\bibliographystyle{IEEEtran}
\bibliography{lexicon}

\end{document}